\documentclass{article}
\usepackage{microtype}
\usepackage{graphicx}
\usepackage{subfigure}
\usepackage{booktabs}
\usepackage{hyperref}

\usepackage[accepted]{icml2020}
\usepackage{multirow}
\usepackage{amsmath}
\usepackage{capt-of}
\usepackage{tabularx}
\usepackage{epsfig}
\usepackage{amssymb}
\usepackage{amsfonts}
\usepackage{scalerel}
\usepackage[inline]{enumitem}
\usepackage{listings}
\usepackage{varwidth}
\usepackage[export]{adjustbox}
\usepackage{tikz}
\usetikzlibrary{tikzmark}
\usepackage{todonotes}
\usepackage{cleveref}

\usepackage{stmaryrd}
\usepackage{bbm}

\newcommand{\tabincell}[2]{\begin{tabular}{@{}#1@{}}#2\end{tabular}}

\newcommand{\eqform}[1]{Equation~(\ref{#1})}

\definecolor{mydarkblue}{rgb}{0,0.08,0.45}
\definecolor{deepblue}{rgb}{0,0,0.5}
\definecolor{officeblue}{RGB}{0,102,204}
\definecolor{deepred}{rgb}{0.6,0,0}
\definecolor{deepgreen}{rgb}{0,0.5,0}
\definecolor{mybrickred}{RGB}{182,50,28}

\definecolor{fillcolor}{RGB}{216,217,252}

\newcommand*\AlgCommentInLine[1]{{\color{deepblue}{$\triangleright$ \textit{#1}}}}

\usepackage{amsmath,amsfonts,bm}

\def\eqref#1{equation~\ref{#1}}

\def\1{\bm{1}}

\DeclareMathAlphabet{\mathsfit}{\encodingdefault}{\sfdefault}{m}{sl}
\SetMathAlphabet{\mathsfit}{bold}{\encodingdefault}{\sfdefault}{bx}{n}

\newcommand{\softmax}{\mathrm{softmax}}

\newcommand\pmlmfull{pseudo-masked language model}
\newcommand\pmlm{\textsc{PMLM}}

\newcommand{\bert}{BERT}
\newcommand{\bertbase}{BERT$_{\textsc{base}}$}

\newcommand{\bartbase}{BART$_{\textsc{base}}$}
\newcommand{\bartlarge}{BART$_{\textsc{large}}$}
\newcommand{\xlnet}{XLNet}
\newcommand{\xlnetbase}{XLNet$_{\textsc{base}}$}

\newcommand{\roberta}{RoBERTa}
\newcommand{\robertabase}{RoBERTa$_{\textsc{base}}$}

\newcommand{\unilmvone}{\textsc{UniLM}}

\newcommand{\vonelarge}{\textsc{UniLM}$_{\textsc{large}}$}
\newcommand{\unilmvtwo}{\textsc{UniLM}v2}
\newcommand{\vtwobase}{\textsc{UniLM}v2$_{\textsc{base}}$}

\newcommand\norelpos{-- rel pos}
\newcommand\norelposfull{-- relative position bias}
\newcommand{\sptk}[1]{\texttt{[#1]}}
\newcommand{\tblidx}[1]{{\small \texttt{[#1]}}}

\icmltitlerunning{\textsc{UniLM}v2: Pseudo-Masked Language Models for Unified Language Model Pre-Training}

\begin{document}

\twocolumn[
\icmltitle{\textsc{UniLM}v2: Pseudo-Masked Language Models for \\ Unified Language Model Pre-Training}

\begin{icmlauthorlist}
\icmlauthor{Hangbo Bao}{msr,hit}
\icmlauthor{Li Dong}{msr}
\icmlauthor{Furu Wei}{msr}
\icmlauthor{Wenhui Wang}{msr}
\icmlauthor{Nan Yang}{msr}
\icmlauthor{Xiaodong Liu}{msr}
\icmlauthor{Yu Wang}{msr}
\icmlauthor{Songhao Piao}{hit}
\icmlauthor{Jianfeng Gao}{msr}
\icmlauthor{Ming Zhou}{msr}
\icmlauthor{Hsiao-Wuen Hon}{msr}

\end{icmlauthorlist}

\begin{icmlauthorlist}
\url{https://github.com/microsoft/unilm}
\end{icmlauthorlist}

\icmlaffiliation{msr}{Microsoft Research}
\icmlaffiliation{hit}{Harbin Institute of Technology}
\vskip 0.3in
]



\printAffiliationsAndNotice{
} 

\begin{abstract}
We propose to pre-train a unified language model for both autoencoding and partially autoregressive language modeling tasks using a novel training procedure, referred to as a pseudo-masked language model (PMLM). Given an input text with masked tokens, we rely on conventional masks to learn inter-relations between corrupted tokens and context via autoencoding, and pseudo masks to learn intra-relations between masked spans via partially autoregressive modeling. With well-designed position embeddings and self-attention masks, the context encodings are reused to avoid redundant computation. Moreover, conventional masks used for autoencoding provide global masking information, so that all the position embeddings are accessible in partially autoregressive language modeling. In addition, the two tasks pre-train a unified language model as a bidirectional encoder and a sequence-to-sequence decoder, respectively. Our experiments show that the unified language models pre-trained using PMLM achieve new state-of-the-art results on a wide range of natural language understanding and generation tasks across several widely used benchmarks.
\end{abstract}

\section{Introduction}

Language model (LM) pre-training on large-scale text corpora has substantially advanced the state of the art across a variety of natural language processing tasks~\cite{elmo,gpt,bert,unilm,roberta,xlnet,bart,albert,t5}.
After LM pre-training, the obtained model can be fine-tuned to various downstream tasks.

\begin{figure}[t]
\centering
\includegraphics[width=0.86\linewidth]{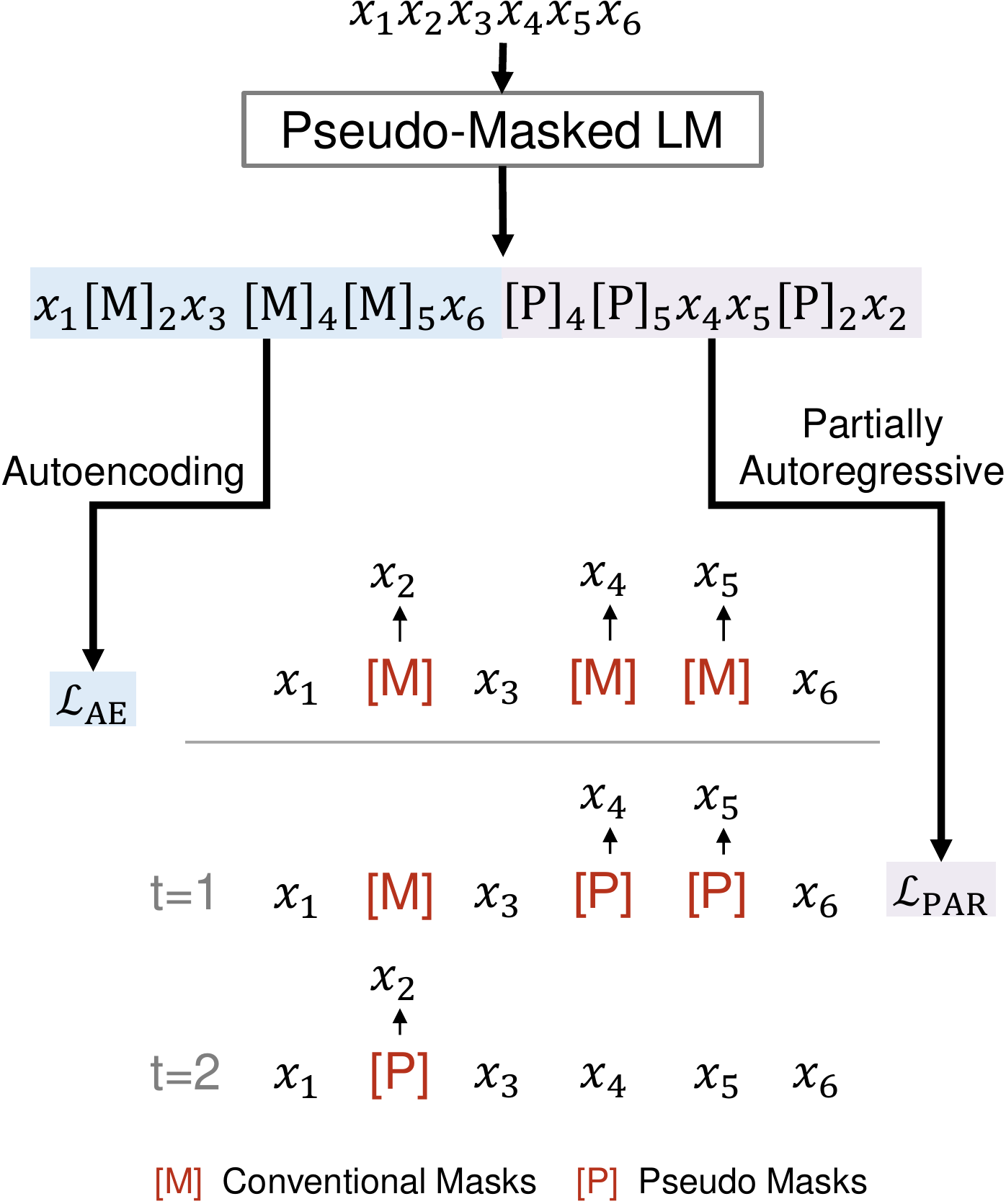}
\caption{
Given input $x_1\cdots x_6$, the tokens $x_2,x_4,x_5$ are masked by the special tokens \sptk{M} and \sptk{P}.
For each example, we jointly train two types of LMs, namely, autoencoding (AE), and partially autoregressive (PAR) masked LMs.
}
\label{fig:intro}
\end{figure}

Two types of language model pre-training objectives are commonly employed to learn contextualized text representations by predicting words conditioned on their context.
The first strand of work relies on autoencoding LMs~\cite{bert,roberta}.
For example, the masked language modeling task used by BERT~\cite{bert} randomly masks some tokens in a text sequence, and then independently recovers the masked tokens by conditioning on the encoding vectors obtained by a bidirectional Transformer~\cite{transformer}.
The second type of pre-training uses autoregressive modeling~\cite{gpt,bart,xlnet,t5}.
Rather than independently predicting words, the probability of a word is dependent on previous predictions.

Inspired by \cite{unilm}, we propose a \pmlmfull{} (\pmlm{}) to jointly pre-train a bidirectional LM for language understanding (e.g., text classification, and question answering) and a sequence-to-sequence LM for language generation (e.g., document summarization, and response generation).
Specifically, the bidirectional model is pre-trained by autoencoding (AE) LMs, and the sequence-to-sequence model is pre-trained by partially autoregressive (PAR) LMs.
As shown in Figure~\ref{fig:intro}, the model parameters are shared in two language modeling tasks, and the encoding results of the given context tokens are reused.
We use the conventional mask \sptk{MASK} (or \sptk{M} for short) to represent the corrupted tokens for AE pre-training.
In order to handle factorization steps of PAR language modeling, we append pseudo masks \sptk{Pseudo} (or \sptk{P} for short) to the input sequence without discarding the original tokens.
With well-designed self-attention masks and position embeddings, the \pmlm{} can perform the two language modeling tasks in one forward pass without redundant computation of context.

The proposed method has the following advantages.
First, the \pmlm{} pre-trains different LMs in a unified manner, which learns both inter-relations between masked tokens and given context (via AE), and intra-relations between masked spans (via PAR).
Moreover, conventional masks used for AE provide global masking information, so that every factorization step of PAR pre-training can access all the position embeddings as in fine-tuning.
Second, the unified pre-training framework learns models for both natural language understanding and generation~\cite{unilm}.
Specifically, the AE-based modeling learns a bidirectional Transformer encoder, and the PAR objective pre-trains a sequence-to-sequence decoder.
Third, the proposed model is computationally efficient in that the AE and PAR modeling can be computed in one forward pass.
Because the encoding results of given context are reused for two language modeling tasks, redundant computation is avoided.
Fourth, PAR language modeling learns token-to-token, token-to-span, and span-to-span relations during pre-training.
By taking spans (i.e., continuous tokens) into consideration, \pmlm{} is encouraged to learn long-distance dependencies by preventing local shortcuts.

We conduct \pmlm{} pre-training on large-scale text corpora.
Then we fine-tune the pre-trained model to a wide range of natural language understanding and generation tasks.
Experimental results show that unified pre-training using \pmlm{} improves performance on various benchmarks.

\section{Preliminary}

\subsection{Backbone Network: Transformer}
\label{sec:transformer}

First, we pack the embeddings of input tokens $\{\textbf{x}_i\}_{i=1}^{|x|}$ together into $\mathbf{H}^0 = [\mathbf{x}_1, \cdots, \mathbf{x}_{|x|}] \in \mathbb{R}^{|x| \times d_h}$.
Then $L$ stacked Transformer~\cite{transformer} blocks compute the encoding vectors via:
\begin{equation}
\mathbf{H}^l = \mathrm{Transformer}_{l}(\mathbf{H}^{l-1}),~l \in [1, L]
\end{equation}
where $L$ is the number of layers.
The hidden vectors of the final layer $\mathbf{H}^L = [\mathbf{h}_1^L, \cdots, \mathbf{h}_{|x|}^L]$ are the contextualized representations of input.
Within each Transformer block, multiple self-attention heads aggregate the output vectors of the previous layer, followed by a fully-connected feed-forward network.

\paragraph{Self-Attention Masks}

The output $\mathbf{A}_l$ of a self-attention head in the $l$-th Transformer layer is:
\begin{align}
\mathbf{Q} &= \mathbf{H}^{l-1} \mathbf{W}_l^Q,~\mathbf{K} = \mathbf{H}^{l-1} \mathbf{W}_l^K \nonumber \\
\mathbf{M}_{ij} &= \begin{cases} 0, &\text{allow to attend} \\ -\infty, &\text{prevent from attending} \end{cases} \label{eq:att:mask} \\
\mathbf{A}_l &= \softmax(\frac{\mathbf{Q} \mathbf{K}^{\intercal}}{ \sqrt{d_k}} + \mathbf{M}) (\mathbf{H}^{l-1} \mathbf{W}_l^V) \nonumber
\end{align}
where parameters $\mathbf{W}_l^Q , \mathbf{W}_l^K , \mathbf{W}_l^V \in \mathbb{R}^{d_h \times d_k}$ project the previous layer's output $\mathbf{H}^{l-1}$ to queries, keys, and values, respectively.
It is worth noting that the mask matrix $\mathbf{M} \in \mathbb{R}^{|x| \times |x|}$ controls whether two tokens can attend each other.

\subsection{Input Representation}
\label{sec:input}

The inputs of language model pre-training are sequences sampled from large-scale text corpora.
We follow the format used by BERT~\cite{bert}.
We add a special start-of-sequence token \sptk{SOS} at the beginning to get the representation of the whole input.
Besides, each text is split into two segments appended with a special end-of-sequence token \sptk{EOS}.
The final input format is ``\sptk{SOS} \texttt{S1} \sptk{EOS} \texttt{S2} \sptk{EOS}'', where the segments \texttt{S1} and \texttt{S2} are contiguous texts.
The vector of an input token is represented by the summation of its token embedding, absolute position embedding, and segment embedding.
All the embedding vectors are obtained by lookup in learnable matrices.

\section{Unified Language Model Pre-Training}

\begin{figure*}[t]
\centering
\includegraphics[width=0.98
\linewidth]{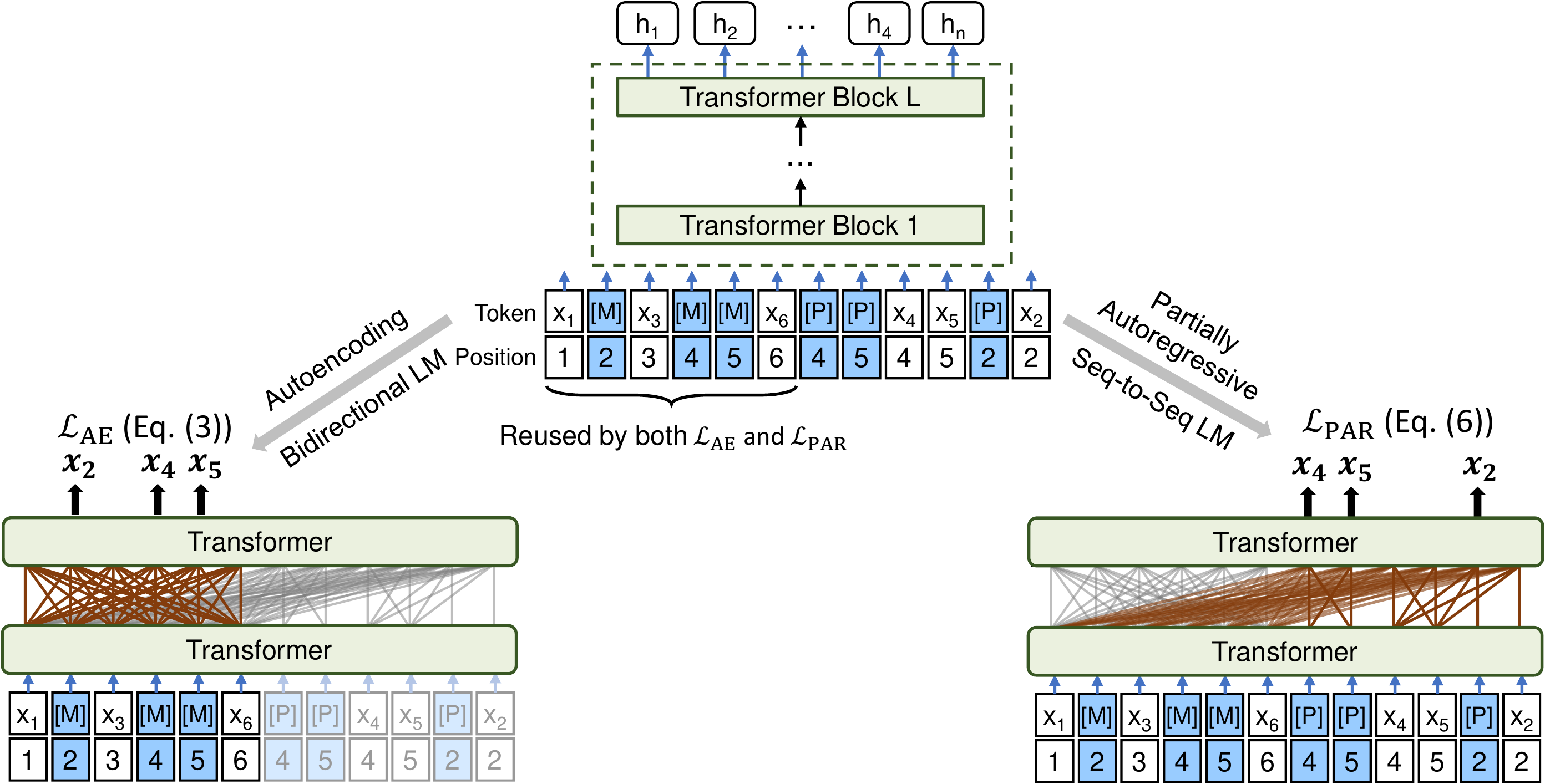}
\caption{Overview of \pmlm{} pre-training. The model parameters are shared across the LM objectives. The bidirectional LM is trained by autoencoding MLM, and the sequence-to-sequence (Seq-to-Seq) LM is trained by partially autoregressive MLM. We use different self-attention masks to control the access to context for each word token.}
\label{fig:overview}
\end{figure*}

We propose a \pmlmfull{} (\pmlm{}) to jointly pre-train both autoencoding (Section~\ref{sec:ae}) and partially autoregressive (Section~\ref{sec:par}) LMs.
As shown in Figure~\ref{fig:overview}, \pmlm{} reuses the encoding results of the same example to jointly pre-train both modeling methods by pseudo masking (Section~\ref{sec:pseudo}).

\subsection{Pre-Training Tasks}
\label{sec:pretrain:tasks}

\begin{table*}[t]
\centering
\begin{tabular}{l l l}
\toprule
& \textbf{Factorization Order} & \textbf{Probability of Masked Tokens} \\
\midrule
\tabincell{l}{Autoencoding (e.g., BERT, and our work)} & $-$ & $p(x_2 | x_{\setminus \{2,4,5\}}) p(x_3 | x_{\setminus \{2,4,5\}}) p(x_5 | x_{\setminus \{2,4,5\}})$ \\ \cmidrule{2-3}
\tabincell{l}{Autoregressive (e.g., GPT, and XLNet)} & \tabincell{l}{$2 \rightarrow 4 \rightarrow 5$ \\ $5 \rightarrow 4 \rightarrow 2$ } & \tabincell{l}{$p(x_2 | x_{\setminus \{2,4,5\}}) p(x_4 | x_{\setminus \{4,5\}}) p(x_5 | x_{\setminus \{5\}})$ \\ $p(x_5 | x_{\setminus \{2,4,5\}}) p(x_4 | x_{\setminus \{2,4\}}) p(x_2 | x_{\setminus \{2\}})$} \\ \cmidrule{2-3}
\tabincell{l}{Partially Autoregressive (our work)} & \tabincell{l}{$2 \rightarrow 4,5$ \\ $4,5 \rightarrow 2$ } & \tabincell{l}{$p(x_2 | x_{\setminus \{2,4,5\}}) p(x_4 | x_{\setminus \{4,5\}}) p(x_5 | x_{\setminus \{4,5\}})$ \\ $p(x_4 | x_{\setminus \{2,4,5\}}) p(x_5 | x_{\setminus \{2,4,5\}}) p(x_2 | x_{\setminus \{2\}})$} \\ \bottomrule
\end{tabular}
\caption{Given input $x = x_1 \cdots x_6$, the tokens $x_2,x_4,x_5$ are masked. We compare how to compute $p(x_2,x_4,x_5 | x_{\setminus \{2,4,5\}})$ with different factorization orders for autoencoding, autoregressive, and partially autoregressive masked language models.}
\label{tbl:compare:ae:ar:par}
\end{table*}

We use the masked language modeling (MLM; \citealt{bert}) task to pre-train a Transformer network, which is also known as the cloze task~\cite{taylor1953cloze}.
For a given input, we randomly substitute tokens with a special token \sptk{MASK} (or \sptk{M} for short). The training objective is to recover them by conditioning on the output hidden states of Transformer.

As shown in Table~\ref{tbl:compare:ae:ar:par}, we categorize MLMs into autoencoding, autoregressive, and partially autoregressive.
Their main difference is how the probability of masked tokens is factorized.
In our work, we leverage autoencoding (AE) and partially autoregressive (PAR) modeling for pre-training, which is formally described as follows.
It is worth noting that the masked positions are the same for both AE and PAR modeling, but the probability factorization is different.

\subsubsection{Autoencoding Modeling}
\label{sec:ae}

The autoencoding method independently predicts the tokens by conditioning on context, which is the same as BERT.
Given original input $x=x_1 \cdots x_{|x|}$ and the positions of masks $M=\{m_1 , \cdots , m_{|M|}\}$, the probability of masked tokens is computed by $\prod_{ m \in M }{ p( x_m |x_{\setminus M} ) }$, where $x_M=\{x_m\}_{m \in M}$, $\setminus$ is set minus, $x_{\setminus M}$ means all input tokens except the ones that are in $M$.
The autoencoding pre-training loss is defined as:
\begin{align}
\mathcal{L}_{\text{AE}} = -{\sum_{x\in \mathcal{D}}{\log{\prod_{ m \in M }{ p( x_m |x_{\setminus M} ) }}}} \label{eq:objective:ae}
\end{align}
where $\mathcal{D}$ is the training corpus.

\subsubsection{Partially Autoregressive Modeling}
\label{sec:par}

We propose to pre-train partially autoregressive MLMs. In each factorization step, the model can predict one or multiple tokens. Let ${M}=\left< M_1 , \cdots , M_{|{M}|} \right>$ denote factorization order, where $M_i=\{ {m}_1^i , \cdots , {m}_{|M_i|}^i \}$ is the set of mask positions in the $i$-th factorization step.
If all factorization steps only contain one masked token (i.e., $|M_i| = 1$), the modeling becomes autoregressive.
In our work, we enable a factorization step to be a span, which makes the LM partially autoregressive.
The probability of masked tokens is decomposed as:
\begin{align}
p( x_M | x_{\setminus M}) =& \prod_{i=1}^{|{M}|}{ p( x_{M_i} | x_{\setminus M_{\ge i} } ) } \\
=& \prod_{i=1}^{|{M}|}{\prod_{ m \in M_i }{ p( x_m | x_{\setminus M_{\ge i}) }}} \label{eq:par}
\end{align}
where $x_{M_i}=\{x_m\}_{m \in M_i}$, and $M_{\ge i}=\bigcup_{j \ge i}{M_j}$.
The partially autoregressive pre-training loss is defined as:
\begin{align}
\mathcal{L}_{\text{PAR}} = -{\sum_{x\in \mathcal{D}}{\mathbb{E}_{M}{\log p\left( x_M | x_{\setminus M}\right)}}} \label{eq:objective:par}
\end{align}
where $\mathbb{E}_{M}$ is the expectation over the factorization distribution.
During pre-training, we randomly sample one factorization order $M$ for each input text~\cite{xlnet}, rather than computing the exact expectation.

\begin{algorithm}[tb]
\centering
\small
\caption{Blockwise Masking}
\label{alg:mask}
\begin{algorithmic}
\STATE {\bfseries Input} $x=x_1 \cdots x_{|x|}$: Input sequence
\STATE {\bfseries Output} ${M}=\left< M_1 , \cdots , M_{|{M}|} \right>$: Masked positions

\STATE {$M \gets \left< \right>$}
\REPEAT{}
\STATE{$p \gets$ \textsf{rand\_int(}$1 , |x|$\textsf{)}}\hfill\AlgCommentInLine{Randomly sample an index}
\STATE{$l \gets$ \textsf{rand\_int(}$2 , 6$\textsf{)}~~\textbf{if} \textsf{rand()} $<$ 0.4 \textbf{else} 1}
\STATE{\textbf{if}~$x_p, \cdots, x_{p+l-1}$ has not been masked~\textbf{then}}
\STATE{~~~~$M\textsf{.append(}\{m\}_{m=p}^{p+l-1}\textsf{)}$}
\UNTIL{$\sum_{j=1}^{|M|}{|M_j|} \ge 0.15|x|$}\hfill\AlgCommentInLine{Masking ratio is $15\%$}
\STATE{\textbf{return} $M$}
\end{algorithmic}
\end{algorithm}

\paragraph{Blockwise Masking and Factorization}
Given input sequence $x$, the masking policy uniformly produces a factorization order ${M}=\left< M_1 , \cdots , M_{|{M}|} \right>$ for \eqform{eq:objective:par}.
For the $i$-th factorization step, the masked position set $M_i$ contains one token, or a continuous text span~\cite{spanbert}.
As described in Algorithm~\ref{alg:mask}, we randomly sample $15\%$ of the original tokens as masked tokens.
Among them, $40\%$ of the time we mask a $n$-gram block, and $60\%$ of the time we mask a token. We then construct a factorization step with the set of masked positions.
We repeat the above process until enough masked tokens are sampled.
The randomly sampled factorization orders are similar to permutation-based language modeling used by XLNet~\cite{xlnet}. However, XLNet only emits predictions one by one (i.e., autoregressive).
In contrast, we can generate one token, or a text span at each factorization step (i.e., partially autoregressive).

\subsection{Pseudo-Masked LM}
\label{sec:pseudo}

\begin{figure}[t]
\centering
\includegraphics[width=0.9\linewidth]{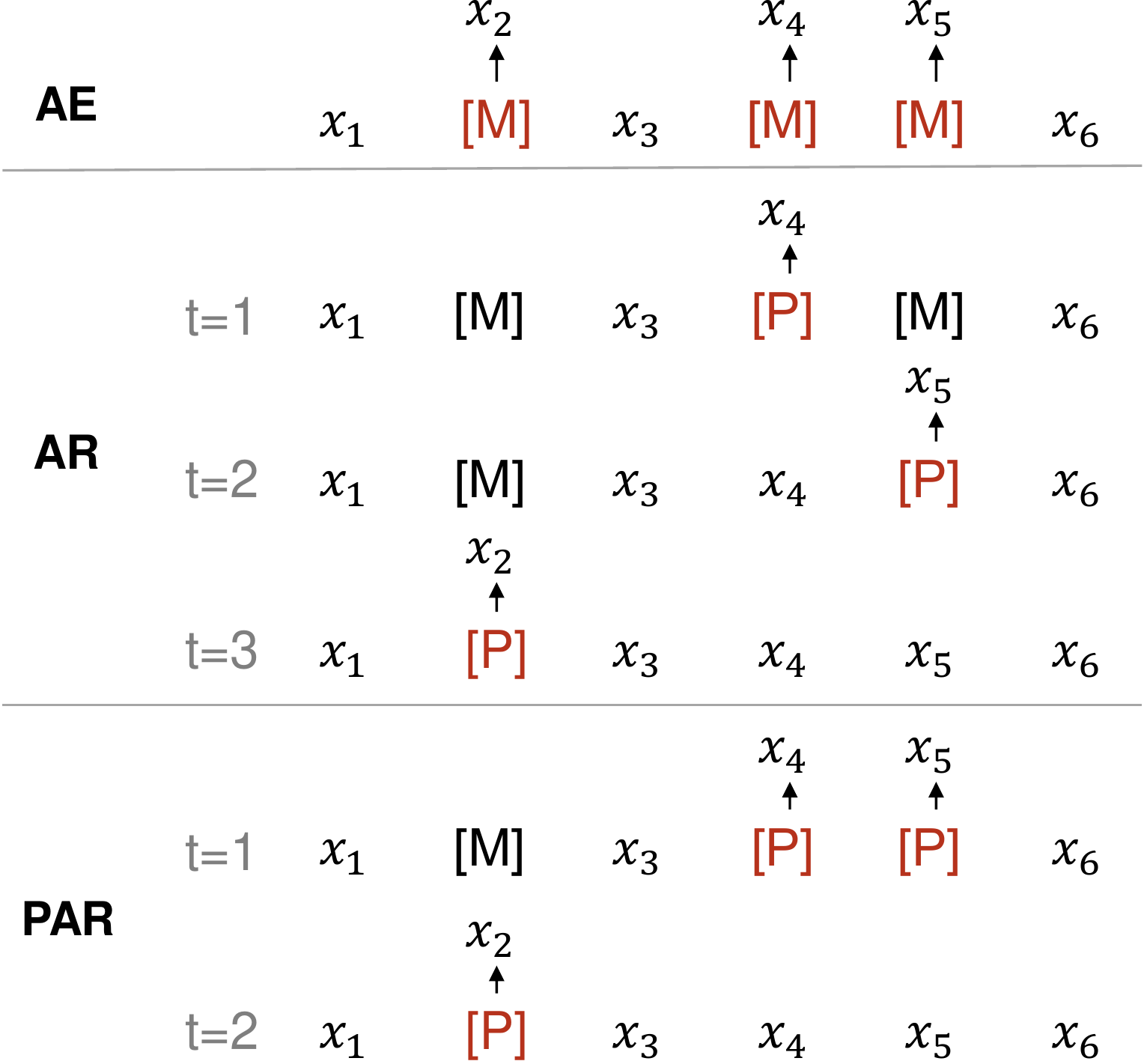}
\caption{
Comparisons between autoencoding (AE), autoregressive (AR), and partially autoregressive (PAR) masked language models.
In the example $x=x_1\cdots x_6$, the tokens $x_2,x_4,x_5$ are masked by the special tokens \sptk{M} and \sptk{P}.
}
\label{fig:ae_ar_par}
\end{figure}

\eqform{eq:par} indicates that factorization steps of partially autoregressive language modeling are conditioned on different context.
So if masked language models~\cite{bert} are directly used, we have to construct a new cloze instance (as shown in Figure~\ref{fig:ae_ar_par}) for each factorization step, which renders partially autoregressive pre-training infeasible.
We propose a new training procedure, named as \pmlmfull{} (\pmlm{}), to overcome the issue.

\begin{figure}[t]
\centering
\includegraphics[width=0.98\linewidth]{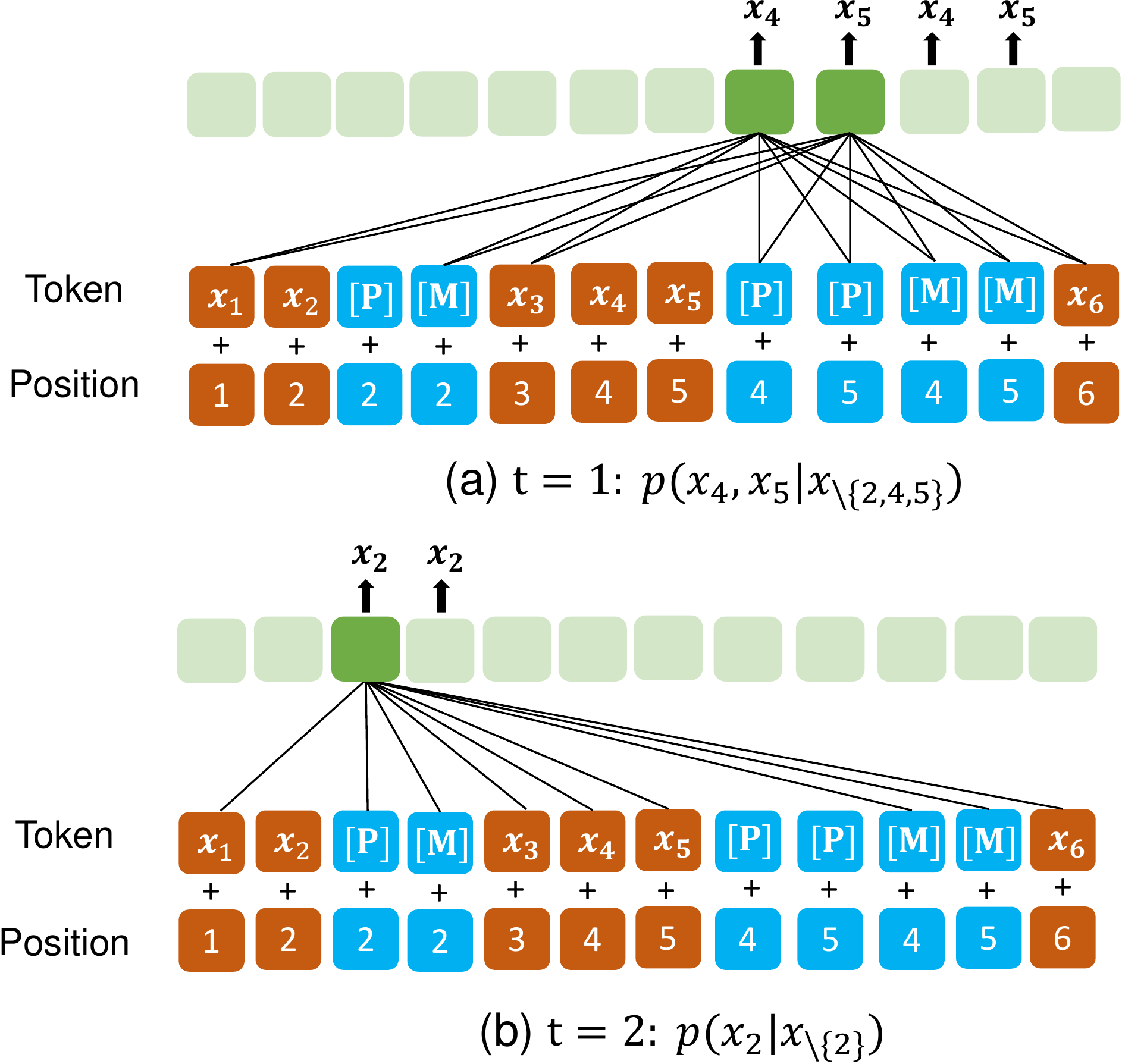}
\caption{
Example of the factorization steps $4,5 \rightarrow 2$.
The masks \sptk{P} and \sptk{M} are assigned with the same position embeddings as the corresponding tokens.
Different context is used to compute the hidden states for the pseudo masks of $x_4,x_5$ and $x_2$.
}
\label{fig:pmlm}
\end{figure}

For the last example in Table~\ref{tbl:compare:ae:ar:par}, Figure~\ref{fig:pmlm} shows how the \pmlm{} conducts partially autoregressive predictions.
Rather than replacing the tokens with masks as in vanilla MLMs, we keep all original input tokens unchanged and append pseudo masks to the input sequence.
For each masked token, we insert a \sptk{Pseudo} (or \sptk{P} for short) token with the same position embedding of the corresponding token. The top-layer hidden states of \sptk{P} tokens are fed into a softmax classifier for MLM predictions.
Notice that positional information in Transformer is encoded by (absolute) position embeddings, while the model components are order-agnostic. In other words, no matter where a token appears in the input sequence, the position of the token is only determined by its position embedding. So we can assign the same position embedding to two tokens, and Transformer treats both of the tokens as if they have the same position.

Vanilla MLMs allow all tokens to attend to each other, while \pmlm{} controls accessible context for each token according to the factorization order.
As shown in Figure~\ref{fig:pmlm}, the example's factorization order is $4,5 \rightarrow 2$.
When we compute $p(x_4,x_5 | x_{\setminus \{2,4,5\}})$, only $x_1,x_3,x_6$ and the pseudo masks of $x_4,x_5$ are conditioned on. The original tokens of $x_4,x_5$ are masked to avoid information leakage, while their pseudo tokens \sptk{P} are used as placeholders for MLM predictions.
In the second step, the tokens $x_1,x_3,x_4,x_5,x_6$ and the pseudo mask of $x_2$ are conditioned on to compute $p(x_2 | x_{\setminus \{2\}})$. Unlike in the first step, the original tokens of $x_4,x_5$ are used for the prediction.

\begin{figure}[t]
\centering
\includegraphics[width=0.76\linewidth]{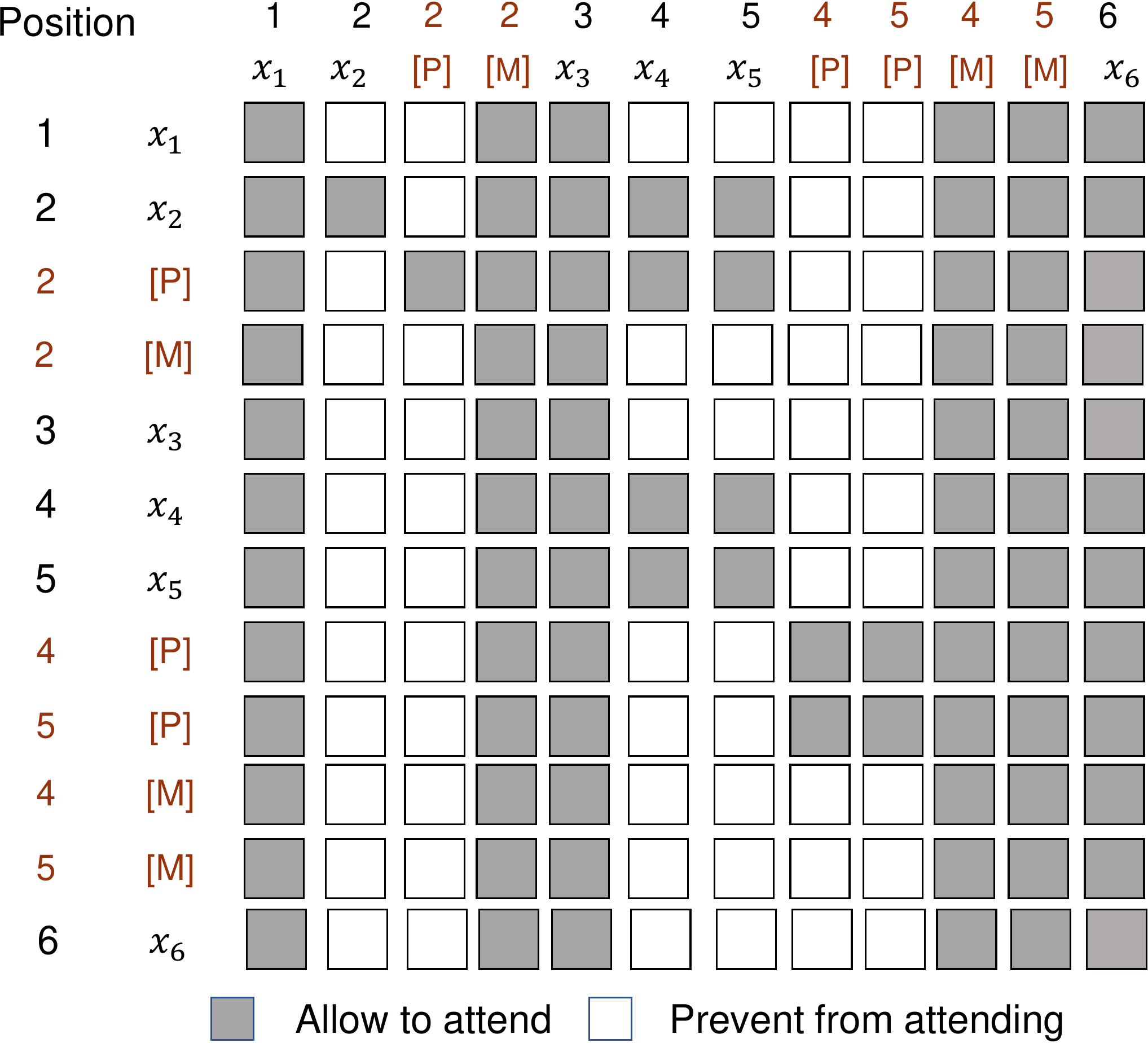}
\caption{
Self-attention mask of the factorization steps $4,5 \rightarrow 2$.
Both conventional masks \sptk{M} and given context ($x_1,x_3,x_6$) can be attended by all the tokens.
}
\label{fig:att:mask}
\end{figure}

Self-attention masks (as described in Section~\ref{sec:transformer}) are used to control what context a token can attend to when computing its contextualized representation.
Figure~\ref{fig:att:mask} shows the self-attention mask matrix used for the example of Figure~\ref{fig:pmlm}.
The self-attention mask matrix is designed in order to avoid two kinds of information leakage.
The first type is \textit{explicit leakage}, i.e., the masked token can be directly accessed by its pseudo token, which renders the LM prediction trivial. So pseudo tokens \sptk{P} are not allowed to attend to the content of ``themselves'' in a \pmlm{}.
The second type is \textit{implicit leakage}, which implicitly leaks prediction information by multi-step attention propagations.
For example, as shown in Figure~\ref{fig:att:mask}, if the context token $x_6$ has access to $x_4$, there is a connected attention flow ``$x_4$'s pseudo mask token $\rightarrow x_6 \rightarrow x_4$'', which eases the prediction of $x_4$.
As a result, for each token, we mask the attentions to the tokens that are predicted in the future factorization steps.

\begin{table*}[t]
\centering
\begin{minipage}{2.48in}
\centering
\begin{tabular}{@{\hskip1pt}l @{\hskip2pt} @{\hskip2pt}c@{\hskip2pt} @{\hskip2pt}c@{ \hskip2pt} @{\hskip2pt}c@{\hskip2pt} @{\hskip2pt}c@{\hskip2pt} @{\hskip2pt}c@{\hskip1pt}}
\toprule
\multirow{2}{*}{\bf Model} & \multicolumn{2}{l}{\bf SQuAD v1.1} &\multicolumn{2}{l}{\bf SQuAD v2.0} \\
& F1 & EM & F1 & EM  \\
\midrule
\bert{} & 88.5 & 80.8 & 76.3 & 73.7 \\
\xlnet{} & - & - & - & 80.2 \\
\roberta{} & 91.5 & 84.6 & 83.7 & 80.5 \\
\unilmvtwo{} & \textbf{93.1} & \textbf{87.1} & \textbf{86.1} & \textbf{83.3} \\
~~\norelpos{} & {93.0} & {86.7} & {85.2} & {82.4} \\
\bottomrule
\end{tabular}
\caption{
Results of \textsc{base}-size pre-trained models on the SQuAD v1.1/v2.0 development sets.
We report F1 and exact match (EM) scores.
Results of \unilmvtwo{} are averaged over five runs.
``\norelpos{}'' is the model without relative position bias.
}
\label{tbl:squad:base}
\end{minipage}
\hfill
\begin{minipage}{4.2in}
\centering
\begin{tabular}{@{\hskip1pt}l@{\hskip2pt} @{\hskip2pt}c@{\hskip3pt} @{\hskip3pt}c@{ \hskip2pt} @{\hskip2pt}c@{\hskip4pt} @{\hskip4pt}c@{\hskip4pt} @{\hskip4pt}c@{\hskip4pt} @{\hskip4pt}c@{\hskip4pt} @{\hskip4pt}c @{\hskip4pt} @{\hskip4pt} c@{\hskip4pt}  @{\hskip4pt} c@{\hskip1pt}}
\toprule
\multirow{2}{*}{\bf Model} & \textbf{MNLI} & \textbf{SST-2} & \textbf{MRPC} & \textbf{RTE}  & \textbf{QNLI} & \textbf{QQP}  & \textbf{STS}  & \textbf{CoLA} \\
                          &      Acc      &      Acc       &      Acc      &      Acc      &      Acc      &      Acc      &      PCC      &      MCC      \\ \midrule
\bert{}                    &     84.5      &      93.2      &     87.3      &     68.6      &     91.7      &     91.3      &     89.5      &     58.9      \\
\xlnet{}                   &     86.8      &      94.7      &     88.2      &     74.0      &     91.7      &     91.4      &     89.5      &     60.2      \\
\roberta{}                 &     87.6      &      94.8      &     90.2      &  78.7 &     92.8      & \textbf{91.9} & \textbf{91.2} &     63.6      \\
\unilmvtwo{}         & \textbf{88.5} & \textbf{95.1}  & \textbf{91.8} & \textbf{81.3}      & \textbf{93.5} & 91.7 & 91.0 & \textbf{65.2} \\
~~\norelpos{}         & {88.4} & {95.0}  & {91.2} &     78.1      & {93.4} &     91.8      & \textbf{91.2} & {63.8} \\
\bottomrule
\end{tabular}
\caption{
Results of \textsc{base}-size models on the development set of the GLUE benchmark.
We report Matthews correlation coefficient (MCC) for CoLA, Pearson correlation coefficient (PCC) for STS, and accuracy (Acc) for the rest.
Metrics of \unilmvtwo{} are averaged over five runs for the tasks.
``\norelpos{}'' is the ablation model without relative position bias.
}
\label{tbl:glue:base}
\end{minipage}
\end{table*}

\subsection{Unified Pre-Training}

As shown in Figure~\ref{fig:overview}, we jointly pre-train bidirectional and sequence-to-sequence LMs with the same input text and masked positions.
Both the special tokens \sptk{M} and \sptk{P} emit predicted tokens.
The training objective is to maximize the likelihood of correct tokens, which considers two types of LMs (i.e., autoencoding, and partially autoregressive) in one example.
The loss is computed via:
\begin{align}
\mathcal{L} = \mathcal{L}_{\text{AE}} + \mathcal{L}_{\text{PAR}}
\end{align}
where $\mathcal{L}_{\text{AE}}, \mathcal{L}_{\text{PAR}}$ are defined as in \eqform{eq:objective:ae}, and \eqform{eq:objective:par} respectively.
The proposed method sufficiently reuses the computed hidden states for both LM objectives.
In addition, experiments in Section~\ref{sec:effect:obj} show that the pre-training tasks are complementary to each other, as they capture both inter- (i.e., between given context and masked tokens) and intra- (i.e., among masked tokens) relations of the input tokens.

\subsection{Fine-tuning on NLU and NLG Tasks}
\label{sec:fine-tuning}

Following~\cite{unilm}, we fine-tune the pre-trained \pmlm{} (with additional task-specific layers if necessary) to both natural language understanding (NLU) and natural language generation (NLG) tasks.

For NLU tasks, we fine-tune \pmlm{} as a bidirectional Transformer encoder, like BERT.
Let us take text classification as an example.
Similar to the text format described in Section~\ref{sec:input}, the input is ``\sptk{SOS} \texttt{TEXT} \sptk{EOS}''.
We use the encoding vector of \sptk{SOS} as the representation of input, and then feed it to a randomly initialized softmax classifier (i.e., the task-specific output layer).
We maximize the likelihood of the labeled training data by updating the parameters of the pre-trained \pmlm{} and the added softmax classifier.

For sequence-to-sequence generation tasks, the example is concatenated as ``\sptk{SOS} \texttt{SRC} \sptk{EOS} \texttt{TGT} \sptk{EOS}'', where \texttt{SRC} and \texttt{TGT} are source and target sequences, respectively.
The fine-tuning procedure is similar to pre-training as in Section~\ref{sec:pseudo}.
For a source sequence, the dependencies between the tokens are bidirectional, i.e., all the source tokens can attend to each other.
In contrast, the target sequence is produced in an autoregressive manner.
So we append a pseudo mask \sptk{P} for each target token, and use self-attention masks to perform autoregressive generation.
The fine-tuning objective is to maximize the likelihood of the target sequence given source input.
It is worth noting that \sptk{EOS} is used to mark the end of the target sequence.
Once \sptk{EOS} is emitted, we terminate the generation process of the target sequence.
During decoding, we use beam search to generate the target tokens one by one~\cite{unilm}.

\begin{table*}[t]
\centering
\begin{tabular}{lllccclccc}
\toprule
\multirow{2}{*}{\bf Model}                          & \multirow{2}{*}{\bf \#Param} & & \multicolumn{3}{c}{\bf CNN/DailyMail} & & \multicolumn{3}{c}{\bf XSum} \\
&                              & & RG-1  & RG-2  &      RG-L      & & RG-1  & RG-2  &     RG-L     \\ \midrule
\multicolumn{8}{l}{\textit{Without pre-training}} & & \\
\textsc{Lead}-3 &                                                         & & 40.42 & 17.62 &     36.67      & & 16.30 & 1.60  &    11.95     \\
\textsc{PtrNet}~\cite{see-2017-get} &                                      & & 39.53 & 17.28 &     36.38      & & 28.10 & 8.02  &    21.72     \\ \midrule
\multicolumn{8}{l}{\textit{Fine-tuning \textsc{large}-size pre-trained models}} & & \\
\vonelarge{}~\cite{unilm}                                        & 340M                         & & 43.08 & 20.43 &     40.34      & &   -   &   -   &      -       \\
\bartlarge{}~\cite{bart}                                        & 400M                         & & 44.16 & 21.28 &     40.90      & & 45.14 & 22.27 &    37.25     \\
{T5$_{\textsc{11B}}$}~\cite{t5}                               & 11B                          & & 43.52 & 21.55 & 40.69 & &   -   &   -   &      -       \\  \midrule
\multicolumn{8}{l}{\textit{Fine-tuning \textsc{base}-size pre-trained models}} & & \\
{MASS$_{\textsc{base}}$}~\cite{mass}                            & 123M                         & & 42.12 & 19.50 &     39.01      & & 39.75 & 17.24 &    31.95     \\
\textsc{BERTSumAbs}~\cite{bertsum}                                 & 156M                         & & 41.72 & 19.39 &     38.76      & & 38.76 & 16.33 &    31.15     \\
{T5$_{\textsc{base}}$}~\cite{t5}                              & 220M                         & &   42.05 & 20.34 & 39.40        & &   -   &   -   &      -       \\
\vtwobase{}                                         & 110M                         & &   {43.16} & {20.42} & {40.14} & & \textbf{44.00} & \textbf{21.11} & \textbf{36.08}     \\
~~\norelposfull{}                                         & 110M                         & &   \textbf{43.45} & \textbf{20.71} & \textbf{40.49} & & 43.69 & 20.71 &    35.73     \\
\bottomrule
\end{tabular}
\caption{
Abstractive summarization results on CNN/DailyMail and XSum.
The evaluation metric is the F1 version of ROUGE (RG) scores.
We also present the number of parameters (\#Param) for the methods using pre-trained models.
}
\label{tbl:summ:base}
\end{table*}

\begin{table}[t]
\centering
\begin{tabular}{l l c c c}
\toprule
    & \textbf{\#Param} & \textbf{BLEU-4} & \textbf{MTR}   & \textbf{RG-L}  \\ \midrule
\multicolumn{2}{l}{\cite{du-qg-2018}}    & 15.16           & 19.12          & -              \\
\multicolumn{2}{l}{\cite{zhang-qg-2019}} & 18.37           & 22.65          & 46.68          \\
\vonelarge{} & 340M        & 22.78           & 25.49          & 51.57          \\
\vtwobase{} & 110M         & {24.43} & \textbf{26.34} & {51.97} \\
~~\norelpos{} & 110M          & \textbf{24.70}  & {26.33} & \textbf{52.13} \\
\midrule
\multicolumn{2}{l}{\cite{zhao-qg-2018}}  & 16.38           & 20.25          & 44.48          \\
\multicolumn{2}{l}{\cite{zhang-qg-2019}} & 20.76           & 24.20          & 48.91          \\
\vonelarge{} & 340M        & 24.32           & 26.10          & 52.69          \\
\vtwobase{} & 110M          & {26.29} & \textbf{27.16} & \textbf{53.22} \\
~~\norelpos{} & 110M          & \textbf{26.30}  & {27.09} & {53.19} \\
\bottomrule
\end{tabular}
\caption{
Results on question generation.
The first block follows the data split in~\cite{du-qg-2018}, while the second block is the same as in~\cite{zhao-qg-2018}.
MTR is short for METEOR, and RG for ROUGE.
``\#Param'' indicates the size of pre-trained models.
``\norelpos{}'' is the model without relative position bias.
}
\label{tbl:qg:base}
\end{table}

\section{Experimental Results}

We employ \pmlmfull{} to conduct unified language model pre-training (\unilmvtwo{}), and fine-tuned the model on both natural language understanding (i.e., question answering, the GLUE benchmark) and generation (i.e., abstractive summarization, and question generation) tasks.
Details about hyperparameters and datasets can be found in the supplementary material.
In addition, we conducted ablation studies to compare different choices of pre-training objectives.

\subsection{Pre-Training Setup}
\label{sec:pretrain:setup}

We followed the same model size as \bertbase{}~\cite{bert} for comparison purposes.
Specifically, we used a $12$-layer Transformer with $12$ attention heads.
The hidden size was $768$, and inner hidden size of feed-forward network was $3072$.
The weight matrix of the softmax classifier was tied with the token embedding matrix.
We also add relative position bias~\cite{t5} to attention scores.
The whole model contains about $110$M parameters.

For fair comparisons, we report the major results using similar pre-training datasets and optimization hyperparameters as in \robertabase{}~\cite{roberta}.
We use 160GB text corpora from English Wikipedia, BookCorpus~\cite{bookcorpus}, OpenWebText\footnote{\url{skylion007.github.io/OpenWebTextCorpus}}, CC-News~\cite{roberta}, and Stories~\cite{stories_data}.
We follow the preprocess and the uncased WordPiece~\cite{gnmt} tokenization used in~\cite{bert}.
The vocabulary size was $30,522$. The maximum length of input sequence was $512$. The token masking probability was $15\%$.
Among masked positions, $80\%$ of the time we replaced the token with masks, $10\%$ of the time with a random token, and keeping the original token for the rest.
The block masking (see Algorithm~\ref{alg:mask}) can mask up to $6$-gram for one factorization step in partially autoregressive modeling.
The batch size was set to $7680$.
We used Adam~\cite{adam} with $\beta_1=0.9$, $\beta_2=0.98$, and $\epsilon=$ 1e-6 for optimization.
The peak learning rate was set to 6e-4, with linear warmup over the first $24,000$ steps and linear decay.
The weight decay was $0.01$. The dropout rate was set to $0.1$.
We ran the pre-training procedure for $0.5$ million steps, which took about $20$ days using $64$ Nvidia V100-32GB GPU cards.

\begin{table*}[t]
\centering
\begin{tabular}{lllccccccc}
\toprule
& \multirow{2}{*}{\bf Model} & \multirow{2}{*}{\bf Objective} & \multicolumn{2}{c}{\textbf{SQuAD v1.1}} & \multicolumn{2}{c}{\textbf{SQuAD v2.0}} & \multicolumn{2}{c}{\textbf{MNLI}} & {\bf SST-2} \\
& & & F1 & EM & F1 & EM & m & mm & Acc \\
\midrule
&\bertbase{} & AE & 88.5 & 80.8 & 76.3 & 73.7 & 84.3 & 84.7 & 92.8 \\
&\xlnetbase{} & AR & - & - & 81.0 & 78.2 & 85.6 & 85.1 & \textbf{93.4} \\
&\robertabase{} & AE & 90.6 & - & 79.7 & - & 84.7 & - & 92.7 \\
&\bartbase{} & AR & 90.8 & - & - & - & 83.8 & - & - \\
\midrule
\tblidx{1} &\vtwobase{} & AE+PAR & \textbf{92.0} & \textbf{85.6} & \textbf{83.6} & \textbf{80.9} & \textbf{86.1} & \textbf{86.1} & 93.2  \\
\tblidx{2} &\tblidx{1} \norelposfull & AE+PAR & 91.5 & 85.0 & 81.8 & 78.9 & 85.6 & 85.5 & 93.0 \\
\tblidx{3} &\tblidx{2} -- blockwise factorization & AE+AR & 90.8 & 84.1 & 80.7 & 77.8 & 85.4 & 85.5 & 92.6 \\
\tblidx{4} &\tblidx{2} -- PAR & AE & 91.0 & 84.2 & 81.3 & 78.4 & 84.9 & 85.0 & 92.4 \\
\tblidx{5} &\tblidx{2} -- AE & PAR & 90.7 & 83.9 & 79.9 & 77.0 & 84.9 & 85.2 & 92.5 \\
\tblidx{6} &\tblidx{5} -- blockwise factorization & AR & 89.9 & 82.9 & 79.3 & 76.1 & 84.8 & 85.0 & 92.3 \\
\bottomrule
\end{tabular}
\caption{
Comparisons between the pre-training objectives.
All models are pre-trained over \textsc{Wikipedia} and \textsc{BookCorpus} for one million steps with a batch size of $256$.
Results in the second block are average over five runs for each task.
We report F1 and exact match (EM) scores for SQuAD, and accuracy (Acc) for MNLI and SST-2.
}
\label{tbl:base_1Mstep}
\end{table*}

\subsection{Question Answering}
\label{sec:qa}

Question answering aims at returning answers for the given question and documents.
We conduct experiments on the benchmarks SQuAD v1.1~\cite{squad1} and v2.0~\cite{squad2}.
The model learns to extract answer spans within a passage.
We formulate the task as a natural language understanding problem.
The input is concatenated as ``\sptk{SOS} \texttt{Question} \sptk{EOS} \texttt{Passage} \sptk{EOS}''.
We add a classification layer on the pre-trained \pmlm{}, which predicts whether each token is the start or end position of an answer span by conditioning on the final outputs of Transformer.
For SQuAD v2.0, we use the output vector of \sptk{SOS} to predict whether the instance is unanswerable or not.

The fine-tuning results are presented in Table~\ref{tbl:squad:base}, where we report F1 scores and exact match (EM) scores.
We compare previous \textsc{base}-size models with \pmlm{}.
Notice that the publicly available \bertbase{} checkpoint~\cite{bert} is pre-trained on 13GB corpora with $256$ batch size, while \xlnetbase{} and \robertabase{} are more directly comparable.
The results show that \vtwobase{} achieves better performance than the other models on both SQuAD datasets.

\subsection{GLUE Benchmark}
\label{sec:glue}

The General Language Understanding Evaluation (GLUE) benchmark~\cite{wang2018glue} contains various tasks.
There are two single-sentence classification tasks, i.e., linguistic acceptability (CoLA; \citealt{cola2018}), and sentiment analysis (SST-2; \citealt{sst2013}).
The text similarity (STS; \citealt{sts-b2017}) task is formulated as a regression problem.
The other tasks are pairwise classification tasks, including natural language inference (RTE, MNLI; \citealt{rte1,rte2,rte3,rte5,mnli2017}), question answering (QNLI; \citealt{squad1}), and paraphrase detection (QQP, MRPC; \citealt{mrpc2005}).


Table~\ref{tbl:glue:base} presents the results on GLUE.
We compare \pmlm{} with three strong pre-trained models, i.e., BERT~\cite{bert}, XLNet~\cite{xlnet}, and RoBERTa~\cite{roberta}, in the single task fine-tuning setting.
All the models are in \textsc{base}-size for fair comparisons.
We observe that the proposed \vtwobase{} outperforms both \bertbase{} and \xlnetbase{} across $8$ tasks. Comparing to state-of-the-art pre-trained \robertabase{}, \vtwobase{} obtains the best performance on $6$ out of $8$ tasks, e.g., $88.4$ vs $87.6$ (\robertabase{}) in terms of MNLI accuracy, indicating the effectiveness of our \vtwobase{}. 

\subsection{Abstractive Summarization}

We evaluate the pre-trained \pmlm{} on two abstractive summarization datasets, i.e., XSum~\cite{xsum}, and the non-anonymized version of CNN/DailyMail~\cite{see-2017-get}.
This is a language generation task, where the texts (such as news articles) are shortened to readable summaries that preserve salient information of the original texts.
The pre-trained \pmlm{} is fine-tuned as a sequence-to-sequence model as described in Section~\ref{sec:fine-tuning}.

We report ROUGE scores~\cite{lin-2004-rouge} on the datasets.
Table~\ref{tbl:summ:base} shows two baseline methods that do not rely on pre-training.
\textsc{Lead}-3 uses the first three input sentences as the summary.
\textsc{PtrNet}~\cite{see-2017-get} is a sequence-to-sequence model with pointer networks.
Results indicate that pre-training achieves significant improvements over the baselines.
We also compare \vtwobase{} with state-of-the-art pre-trained models of both \textsc{base}-size and \textsc{large}-size.
We focus on the comparisons in the third block because the models contain similar numbers of parameters.
\textsc{BERTSumAbs}~\cite{bertsum} fine-tunes a BERT encoder that is pre-trained with an autoencoding objective, concatenating with a randomly initialized decoder.
MASS~\cite{mass} and T5~\cite{t5} pre-train encoder-decoder Transformers with masked LM, which relies on the autoregressive pre-training.
Although \pmlm{} has the smallest size, we find that \vtwobase{} outperforms the other \textsc{base}-size pre-trained models on both datasets.

\subsection{Question Generation}

We perform evaluations on question generation~\cite{du-qg-2018}, the task of automatically producing relevant questions that ask for the given answer and context.
The input of the sequence-to-sequence problem is defined as the concatenation of a paragraph and an answer.
We fine-tune the pre-trained \pmlm{} to predict output questions.

As shown in Table~\ref{tbl:qg:base}, we report BLEU~\cite{bleu}, METEOR\cite{meteor}, and ROUGE~\cite{lin-2004-rouge} scores on two different data splits.
Among the compared results, \unilmvone{}~\cite{unilm} is based on pre-trained models, while the other three methods are sequence-to-sequence models enhanced with manual features~\cite{du-qg-2018}, gated self-attention~\cite{zhao-qg-2018}, and reinforcement learning~\cite{zhang-qg-2019}.
Results show that \vtwobase{} achieves better evaluation metrics compared with \vonelarge{} and several baselines.
It is worth noting that \vtwobase{} consists of three times fewer parameters than \vonelarge{}.

\subsection{Effect of Pre-Training Objectives}
\label{sec:effect:obj}

We conduct ablation experiments on using \pmlm{} to implement different pre-training objectives, i.e., autoencoding (AE), autoregressive (AR), partially autoregressive (PAR), and jointly training (AE+AR, and AE+PAR).
The evaluations follow the same settings\footnote{Models were trained for 1M steps with batch size of $256$ over English Wikipedia and BookCorpus~\cite{bookcorpus}. The learning rate of Adam ($\beta_1=0.9$, $\beta_2=0.999$) was set to 1e-4, with linear schedule and warmup over the first $10$K steps.} as in BERT~\cite{bert}, so that the results in Table~\ref{tbl:base_1Mstep} can be directly compared with each other.
Notice that XLNet~\cite{xlnet} is an autoregressive MLM augmented with more advanced relative position embeddings, and long-context memory.

As shown in Table~\ref{tbl:base_1Mstep}, we compare the \pmlm{}-based variants against previous models on question answering (SQuAD; \citealt{squad1,squad2}), natural language inference (MNLI; \citealt{mnli2017}), and sentiment classification (SST-2; \citealt{sst2013}).
First, we ablate relative position bias to better compare with BERT, RoBERTa, and BART.
On text classification (MNLI and SST-2), the PAR-only objective compares favorably with both AE-only and AR-only objectives, which indicates the effectiveness of the proposed PAR modeling.
In comparison, the SQuAD tasks require more precise modeling of spans in order to extract correct answer spans from the input passage, where both AE-only and PAR-only objectives outperform the AR-only objective.
The results indicate that block masking and factorization are important for LM pre-training.
Besides, the settings of jointly training (AE+AR, and AE+PAR) tend to improve the results over using single LM task.
Among the five objectives, AE+PAR performs the best with the help of \pmlm{}, which shows that autoencoding and partially autoregressive modelings are complementary for pre-training.

\section{Conclusion}

We pre-train a unified language model for language understanding and generation by joint learning bidirectional LM (via AE) and sequence-to-sequence LM (via PAR).
We introduce a \pmlmfull{} (\pmlm{}) to efficiently realize the unified pre-training procedure.
\pmlm{} is computationally efficient in that AE and PAR can be computed in one forward pass without redundant computation.
Besides, the two modeling tasks are complementary to each other.
Because conventional masks of AE provide global masking information to PAR, and PAR can learn intra-relations between masked spans.
In addition, the proposed PAR pre-training encourages to learn long-distance dependencies by preventing local shortcuts.
Experimental results show that \pmlm{} improves the end-task results on several language understanding and generation benchmarks.

\bibliography{pmlm}
\bibliographystyle{icml2020}

\clearpage
\appendix

\section{Hyperparameters for Pre-Training}

As shown in Table~\ref{tbl:pretraining_hyperparams}, we present the hyperparameters used for pre-training \vtwobase{}.
We use the same WordPiece~\cite{gnmt} vocabulary and model size as \bertbase{}~\cite{bert}.
We follow the optimization hyperparameters of \robertabase{}~\cite{roberta} for comparisons.

\begin{table}[h]
\centering
\begin{tabular}{lc}
\toprule
Layers & 12 \\
Hidden size & 768 \\
FFN inner hidden size & 3072 \\
Attention heads & 12 \\
Attention head size & 64 \\
Max relative position & 128 \\
Training steps & 0.5M \\
Batch size & 7680 \\
Adam $\epsilon$ & 1e-6 \\
Adam $\beta$ & (0.9, 0.98) \\
Learning rate & 6e-4 \\
Learning rate schedule & Linear \\
Warmup ratio & 0.048 \\
Gradient clipping & 0.0 \\
Dropout & 0.1 \\
Weight decay & 0.01 \\
\bottomrule
\end{tabular}
\caption{
Hyperparameters for pre-training \vtwobase{}.
}
\label{tbl:pretraining_hyperparams}
\end{table}

\section{GLUE Benchmark}

Table~\ref{tbl:glue:datasets} summarizes the datasets used for the General Language Understanding Evaluation (GLUE) benchmark~\cite{wang2018glue}.

\begin{table}[h]
\centering
\begin{tabular}{l l}
\toprule 
\textbf{Dataset} & \textbf{\#Train/\#Dev/\#Test}   \\ \midrule
\multicolumn{2}{l}{\emph{Single-Sentence Classification}} \\
CoLA (Acceptability)&8.5k/1k/1k \\
SST-2 (Sentiment)&67k/872/1.8k \\ \midrule
\multicolumn{2}{l}{\emph{Pairwise Text Classification}} \\
MNLI (NLI)& 393k/20k/20k\\
RTE (NLI) &2.5k/276/3k \\ 
QNLI (NLI)& 105k/5.5k/5.5k\\
WNLI (NLI) &634/71/146\\ 
QQP (Paraphrase)&364k/40k/391k\\ 
MRPC (Paraphrase) &3.7k/408/1.7k\\ \midrule
\multicolumn{2}{l}{\emph{Text Similarity}} \\
STS-B (Similarity) &7k/1.5k/1.4k \\ \bottomrule
\end{tabular}
\caption{Summary of the GLUE benchmark.
}
\label{tbl:glue:datasets}
\end{table}

\section{Hyperparameters for NLU Fine-Tuning}

Table~\ref{tbl:nlu_finetune_hyperparams} reports the hyperparameters used for fine-tuning \vtwobase{} over SQuAD v1.10~\cite{squad1} / v2.0~\cite{squad2}, and the GLUE benchmark~\cite{wang2018glue}.
The hyperparameters are searched on the development sets according to the average performance of five runs.
We use the same hyperparameters for both SQuAD question answering datasets.
Moreover, we list the hyperparameters used for the GLUE datasets in Table~\ref{tbl:nlu_finetune_hyperparams}.

\begin{table}[h]
\centering
\small
\begin{tabular}{@{\hskip3pt}l@{\hskip2pt}c@{\hskip2pt}c@{\hskip2pt}c@{\hskip4pt}}
\toprule
 & \bf SQuAD v1.1/v2.0 & \bf GLUE \\
\midrule
Batch size & 32  & \{16, 32\}\\
Learning rate & 2e-5 & \{5e-6, 1e-5, 1.5e-5, 2e-5, 3e-5\}\\
LR schedule & \multicolumn{2}{c}{Linear} \\
Warmup ratio & 0.1 & \{0.1, 0.2\} \\
Weight decay & 0.01 & \{0.01, 0.1\} \\
Epochs & 4 & \{10, 15\} \\
\bottomrule
\end{tabular}
\caption{
Hyperparameters used for fine-tuning on SQuAD, and GLUE.
}
\label{tbl:nlu_finetune_hyperparams}
\end{table}

\section{Hyperparameters for NLG Fine-Tuning}

As shown in Table~\ref{tbl:nlg_finetune_hyperparams}, we present the hyperparameters used for the natural language generation datasets, i.e., CNN/DailyMail~\cite{see-2017-get}, XSum~\cite{xsum}, and SQuAD question generation~(QG; \citealt{du-qg-2018,zhao-qg-2018}).
The total length is set to $512$ for QG, and $768$ for CNN/DailyMail and XSum.
The maximum output length is set to $160$ for CNN/DailyMail, and $48$ for XSum and QG.
The label smoothing~\cite{label:smoothing} rate is $0.1$.
During decoding, we use beam search to generate the outputs.
Length penalty~\cite{gnmt} is also used to score candidates.

\begin{table}[h]
\centering
\small
\begin{tabular}{lccc}
\toprule
 & \bf CNN/DailyMail & \bf XSum & \bf QG \\
\midrule
\multicolumn{4}{l}{\emph{Fine-Tuning}} \\
Learning rate & 7e-5 & 7e-5 & 2e-5 \\
Batch size & 64 & 64 & 48 \\
Weight decay & \multicolumn{3}{c}{0.01} \\
Epochs & 14 & 14 & 16 \\
Learning rate schedule & \multicolumn{3}{c}{Linear} \\
Warmup ratio & 0.02 & 0.02 & 0.1 \\
Label smoothing & \multicolumn{3}{c}{0.1} \\
Max input length & 608 & 720 & 464 \\
Max output length & 160 & 48 & 48 \\
\midrule
\multicolumn{4}{l}{\emph{Decoding}} \\
Length penalty & 0.7 & 0.6 & 1.3 \\
Beam size & 5 & 5 & 8 \\
\bottomrule
\end{tabular}
\caption{
Hyperparameters used for fine-tuning and decoding on CNN/DailyMail, XSum, and question generation (QG).
}
\label{tbl:nlg_finetune_hyperparams}
\end{table}

\end{document}